**UNIVERSITY OF PATRAS**
**DEPARTMENT OF ELECTRICAL AND COMPUTER ENGINEERING**
**Wire Communications Laboratory**
***Speech and Language Technology Group***
GR-265 00 Rio, Greece - tel:+30.61.991722 - fax:+30.61.991855 - http://slt.wcl.ee.upatras.gr

# A Straightforward Approach to Morphological Analysis and Synthesis

**K. Sgarbas, N. Fakotakis, G. Kokkinakis**

# Foreword

This paper presents a straightforward approach to the problem of morphological processing. Specifically, a DAWG (Directed Acyclic Word Graph) is used to store both the lexicon and the accompanied morphological information, in such a way that both morphological analysis and synthesis can be reduced to a search in the graph. We have used the algorithms described in "Two Algorithms for Incremental Construction of Directed Acyclic Word Graphs" for the creation and update of the structure and the morphological information presented in "A PC-KIMMO-Based Morphological Analysis of Modern Greek" and "A Morphological Description of Modern Greek using the Two-Level Model".

---

If you make references to this paper please inform the authors at: sgarbas@wcl.ee.upatras.gr

# A STRAIGHTFORWARD APPROACH TO MORPHOLOGICAL ANALYSIS AND SYNTHESIS

*Kyriakos N. Sgarbas, Nikos D. Fakotakis, George K. Kokkinakis*
Wire Communications Lab., Electrical & Computer Engineering Dept.,
University of Patras, GR-26500, Greece
sgarbas@wcl.ee.upatras.gr
http://slt.wcl.ee.upatras.gr

## ABSTRACT

In this paper we present a lexicon-based approach to the problem of morphological processing. Full-form words, lemmas and grammatical tags are interconnected in a DAWG. Thus, the process of analysis/synthesis is reduced to a search in the graph, which is very fast and can be performed even if several pieces of information are missing from the input. The contents of the DAWG are updated using an on-line incremental process. The proposed approach is language independent and it does not utilize any morphophonetic rules or any other special linguistic information.

## 1. INTRODUCTION

The morphological processing problem can be summarised in two separate tasks: (a) given a word form, determine its lemma and its grammatical description (this is known as morphological analysis) and (b) given a lemma and a set of intended grammatical features, determine the corresponding inflected form (this is known as morphological synthesis). There is also a third option: given a word in any form and a set of intended grammatical features, change the word to the corresponding inflected form; but this can be reduced easily to a combination of the two aforementioned tasks.

The problem is generally not deterministic and many approaches have been proposed to deal with it. Some of them are language specific using special knowledge of the language in question to obtain the results [5]. Other methods are generally language independent [1,3] but they require the development of some model of the language (e.g. morphophonetic and morphotactic rules).

In this paper we describe a straightforward approach that is language independent, requires no model of the language or any other type of preprogrammed linguistic information and uses a single data structure for both analysis and synthesis. The structure is a specially structured lexicon that contains the lemmas, the inflected forms of the words and their grammatical features stored in a Directed Acyclic Word Graph (DAWG) [4]. Thus the problem of morphological processing is reduced to a search in the graph. The structure has certain similarities to a lexical transducer [2]: it incorporates surface and lexical forms as well as the appropriate morphological information. It differs in the way these forms and information are stored. And since it contains only finite strings, algorithms for incremental construction of DAWGs [5,8] can be applied to update it. These algorithms are fast enough to be used on-line.

The proposed method has been applied to a Greek-language morphological analyser used in the project MITOS[1]. This analyser proved to be much faster (10.000 words/sec) than a two-level morphological analyser with the same number of lemmas that we used before (20 words/sec) [7,8,9].

The method is characterised as "straightforward" because it directly maps inflected word forms to lemmas. The contents of other types of morphological processors and lexicons can be transferred easily to this structure. Its response is instantaneous and it even facilitates content-addressable search, which is useful in case of corrupted or uncertain input data. The only disadvantage of the method is the large size of the structure, which although it is not so serious considering the capabilities of modern computer systems, however it is adequately confronted by the use of DAWGs.

## 2. THE MODEL

### 2.1 The DAWG Data Structure

The DAWG (Directed Acyclic Word Graph) [4] is a very efficient data structure for lexicon representation and fast string matching, with a great variety of applications. It is able to store finite strings of information in a compact way, tak-

[1] Project MITOS (GSRT-EKVAN-2-1.3-102) "Development of a data-mining system for the automatic retrieval of information from financial news"

information in a compact way, taking advantage of common prefixes and suffixes in the strings.

To illustrate this, Fig.1 shows three structures containing the same strings of information (a small lexicon). The strings are stored as directed paths on the graphs. They can be retrieved by traversing the graph from an *initial* node to a *terminal* node, collecting the labels of the links encountered. The left graph in Fig.1 is a trie, which utilizes common string prefixes while representing the lexicon. The other two are DAWGs and they utilize both common string prefixes and suffixes to provide an even more compact representation. There are two types of DAWGs. The middle structure in Fig.1 is a *deterministic* DAWG, meaning that for any of its nodes there are no two departing links that have the same label. This property results to a very efficient search function but adds the need of a *stop* character to mark the end of the string. The right structure in Fig.1 represents a *non-deterministic* DAWG. It lacks the aforementioned property, but it is even more compact. The use of non-deterministic DAWGs is generally the best choice if we plan to built a processor able to function both as an analyser and a synthesiser, but if we need only one function, we can obtain even better response time by using a deterministic one.

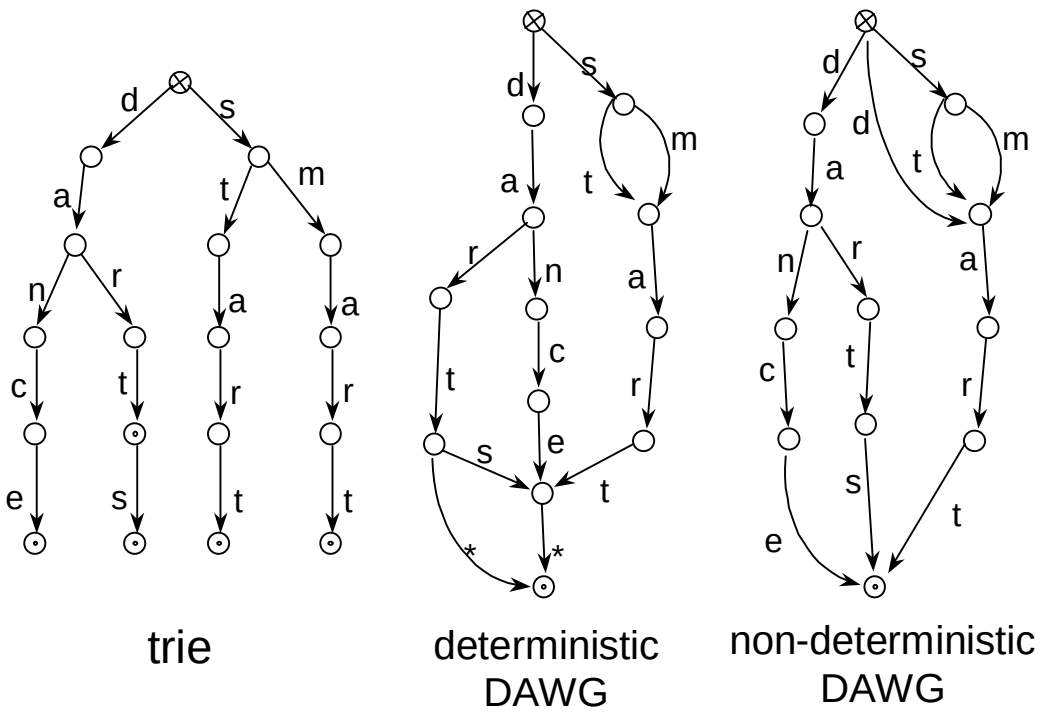

**Figure 1: Graph Representations of Strings**

## 2.2 Representation of Word Entries

Although the example of Fig.1 involves simple words, in the proposed method we used a DAWG to store entries with much more information. Each entry is a string composed by the full (inflected) form of the word, its grammatical features, and its corresponding lemma. Other information that might be useful in some cases, like stem or morphotactic information can be uniquely determined from the lemma, thus it was not included in the representation scheme. If we do need this information we can access it via a look-up table using the lemma as a key.

Figure 2 shows an example entry. It illustrates how the adjective *μονιμότερου* (meaning "of the more permanent") is represented. A string is formed which composes three parts: the full-form word (left), its grammatical features (middle) and its corresponding lemma (right). The lemma *μόνιμος* (permanent) is stored reversed as *ςομινόμ*. This inversion is useful because DAWGs use common links for the first and the last parts of the stored strings, thus reducing the size of the whole structure. And since we obtain better compression for Greek words if we combine their prefixes rather than their suffixes [6,8], the aforementioned inversion helps to produce more compact graphs.

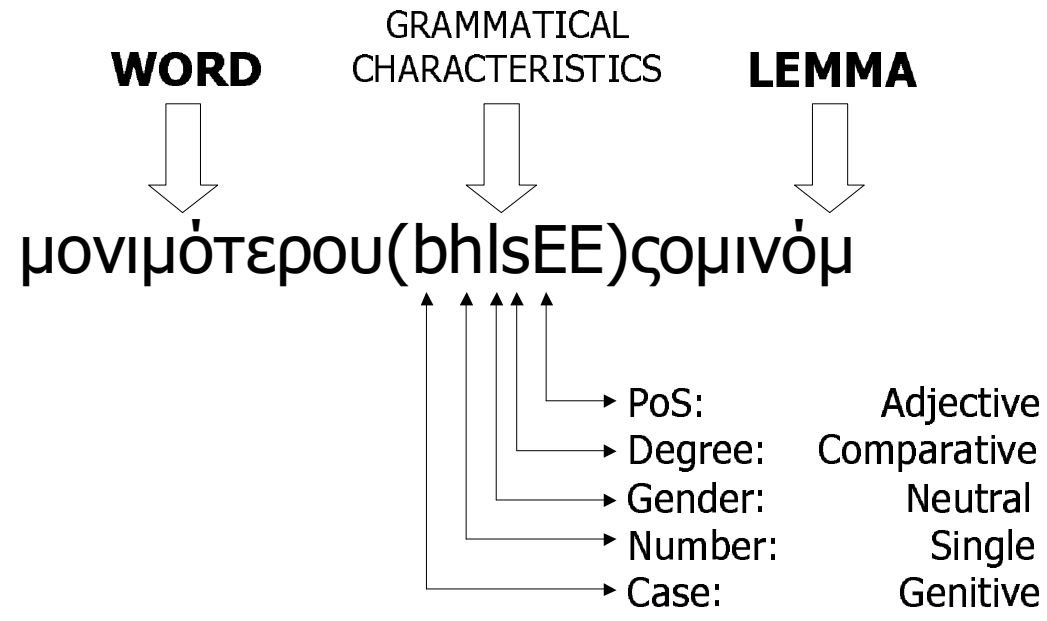

**Figure 2: Example of a Word Entry**

The grammatical features are coded in a substring of characters separated from the word and the lemma by the special symbols "(" and ")". For instance, in the string of Fig.2, character "b" denotes that the word is in genitive case while character "h" suggests single number. The full list of Greek grammatical features and their corresponding symbols are shown in Table 1.

The order of the characters in the substring also affects the size of the DAWG. For this reason the features that are more dependent on the word suffix have been positioned first (close to the suffix), while the ones that are less dependent on the suffix have been put last (close to the lemma). The proper ordering is also represented in the columns of Table 1 from the less suffix-dependent features (leftmost column) to the more suffix-dependent ones (rightmost column).

Also note that in Fig.2 there are two characters "E" representing the part-of-speech information. The first one refers to the POS of the word, while the second one refers to the POS of the lemma. Although normally these two are identical, this is not always the case. In Greek grammar there are adverbs derived from adjectives and participles (which are regarded as separate POS) derived from verbs.

| PART OF SPEECH | TYPE | MODE | VOICE | TENSE | MOOD | DEGREE | GENDER | NUMBER | PERSON | CASE |
|---|---|---|---|---|---|---|---|---|---|---|
| article (A) | definite (ε)<br>indefinite (ζ)<br>prepositional (η) | | | | | | masculine (j)<br>feminine (k)<br>neutral (l) | singular (h)<br>plural (i) | | nominative (a)<br>genitive (b)<br>accusative (c) |
| noun (N) | proper (λ)<br>common (μ) | | | | | | masculine (j)<br>feminine (k)<br>neutral (l) | singular (h)<br>plural (i) | | nominative (a)<br>genitive (b)<br>accusative (c)<br>vocative (d) |
| adjective (E) | absolute numeral (λ)<br>orderly numeral (μ)<br>multiplication numeral (ν)<br>proportional numeral (ξ)<br>collective numeral (ο) | | | | | positive (r)<br>comparative (s)<br>superlative (t) | masculine (j)<br>feminine (k)<br>neutral (l) | singular (h)<br>plural (i) | | nominative (a)<br>genitive (b)<br>accusative (c)<br>vocative (d) |
| pronoun (T) | personal - strong (λ)<br>personal - weak (μ)<br>possessive (ν)<br>reflexive (ξ)<br>indicative (ο)<br>relative (π)<br>interrogative (ρ)<br>indefinite (σ)<br>definite (τ) | | | | | | masculine (j)<br>feminine (k)<br>neutral (l) | singular (h)<br>plural (i) | first (e)<br>second (f)<br>third (g) | nominative (a)<br>genitive (b)<br>accusative (c)<br>vocative (d) |
| verb (V) | transitive (θ)<br>intransitive (ι)<br>with two endings (κ) | active (α)<br>passive (β)<br>middle (γ)<br>neutral (δ) | active (y)<br>passive (z) | present (v)<br>imperfect (w)<br>past (x)<br>pres. perfect (u) | indicative (m)<br>conjunctive (n)<br>imperative (o)<br>infinitive (p)<br>participle (q) | | | singular (h)<br>plural (i) | first (e)<br>second (f)<br>third (g) | |
| participle (M) | transitive (θ)<br>intransitive (ι)<br>with two endings (κ) | active (α)<br>passive (β)<br>middle (γ)<br>neutral (δ) | active (y)<br>passive (z) | present (v)<br>pres. perfect (u) | participle (q) | | masculine (j)<br>feminine (k)<br>neutral (l) | singular (h)<br>plural (i) | | nominative (a)<br>genitive (b)<br>accusative (c)<br>vocative (d) |
| adverb (R) | local (λ)<br>temporal (μ)<br>modal (ν)<br>quantitative (ξ)<br>confirmatory (ο)<br>tentative (π)<br>negative (ρ) | | | | | positive (r)<br>comparative (s)<br>superlative (t) | | | | |
| preposition (P) | | | | | | | | | | |
| conjunction (S) | co-ordinate (λ)<br>separating (μ)<br>oppositional (ν)<br>inferential (ξ)<br>explanatory (ο)<br>special (π)<br>temporal (ρ)<br>causal (σ)<br>conditional (τ)<br>eventual (υ)<br>resulting (φ)<br>tentative (χ)<br>comparative (ψ) | | | | | | | | | |
| particle (I) | protreptic (λ)<br>future (μ)<br>potential (ν)<br>speculative (ξ)<br>volitional (ο)<br>indicative (π)<br>oath (ρ) | | | | | | | | | |
| exclamation (F) | | | | | | | | | | |
| abbreviation (G) | | | | | | | | | | |
| other (X) | | | | | | | | | | |

**Table 1: Coding of Grammatical Features for Greek**

## 2.3 Search for Analysis and Synthesis

To access the information contained in a DAWG, a simple depth-first search is adequate in most of the cases. However, a DAWG is able to facilitate more sophisticated search procedures, like a partial or content addressable search, since it is able to retrieve all entries that match to a given pattern. In our case we use two types of patterns: for morphological analysis we use the pattern ***word(**** to initiate a search from the initial node downwards (where * is a wildcard matching any substring); for morphological synthesis we use the pattern ***(features)lemma*** to initiate a search from the terminal node upwards. The searches are generally non-deterministic, since there are often more than one strings that match the input pattern, but they are very fast.

## 2.4 Update of the Structure

For the update of the structure we have developed two algorithms (one for deterministic and one for non-deterministic DAWGs) that are able to add strings in existing DAWGs on-line. The two algorithms are described in detail in [6,8]. Here we give only an example of operation in Fig.3: The string *stair* is to be added in a deterministic DAWG. The operation is performed in three stages. During Stage 1 the prefix of the new string is matched to the current contents of the original DAWG (Figs.3b-3d). Stage 2 creates the necessary links for the remaining part of the string (Fig.3e). In Stage 3 the new links are combined with already existing ones to obtain a compact representation (Figs.3f-i). The updated DAWG (Fig.3j) contains all the strings of the original one, plus the new string. This operation is quite efficient to be used on-line. Moreover, we have shown [10] that if the original DAWG is minimal (i.e. it has the least possible number of nodes) then the updated DAWG is guaranteed to be also minimal.

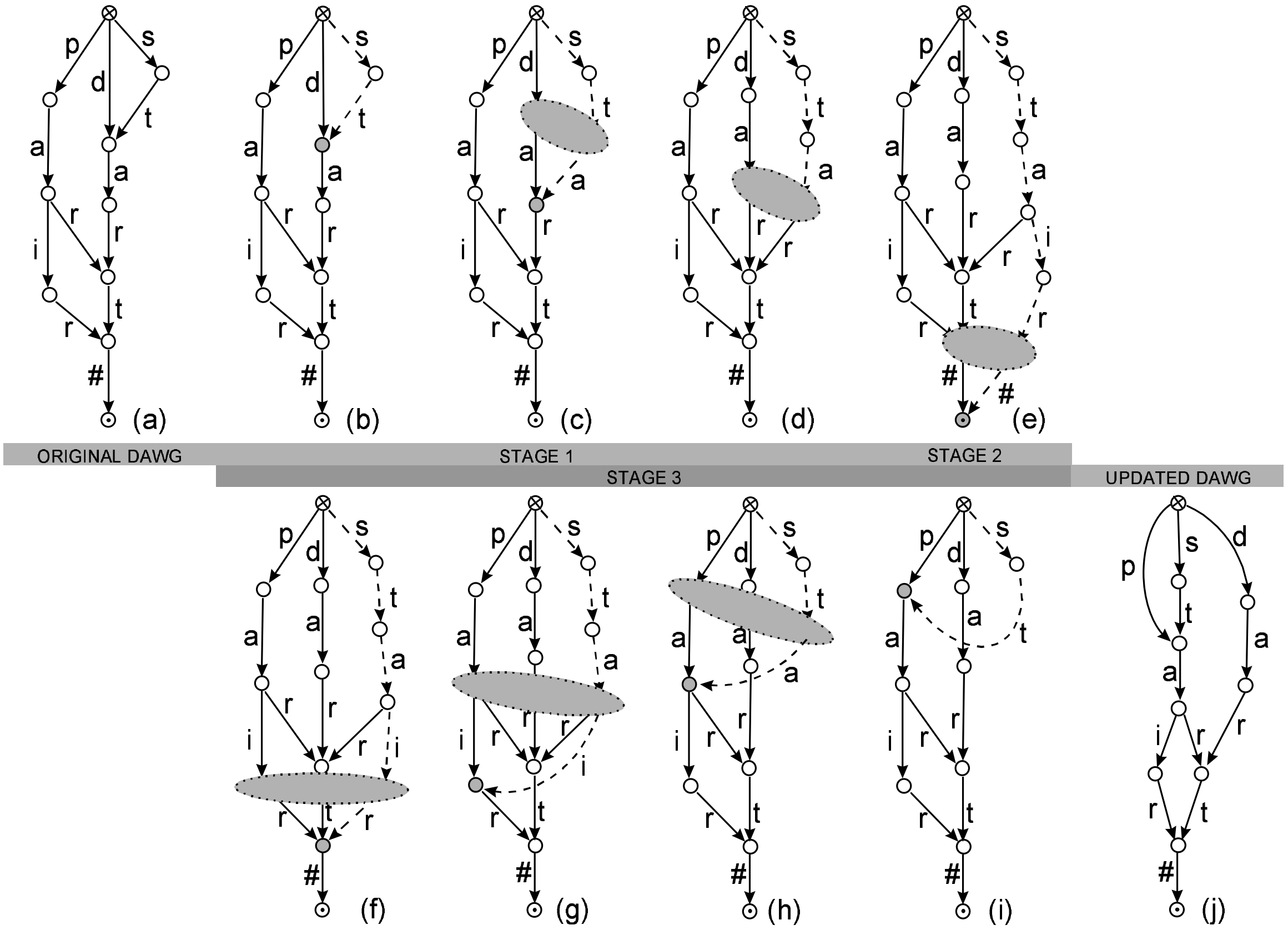

**Figure 3: The String "stair" is Added in a Deterministic DAWG**